\documentclass[letterpaper]{article}
\usepackage{uai2019}
\usepackage[margin=1in]{geometry}

\usepackage{times}
\usepackage{format}

\title{Learning Belief Representations for Imitation Learning in POMDPs}

\author{
  Tanmay Gangwani\\
  Dept. of Computer Science\\
  UIUC\\
  \texttt{gangwan2@uiuc.edu}\\
     \And
   Joel Lehman \\
   Uber AI Labs\\
   San Francisco, CA 94103 \\
   \texttt{joel.lehman@uber.com} \\
     \And
   Qiang Liu\\
   Dept. of Computer Science\\
   UT Austin \\
   \texttt{lqiang@cs.utexas.edu} \\
   \And
   Jian Peng \\
   Dept. of Computer Science \\
   UIUC \\
   \texttt{jianpeng@uiuc.edu} \\
}

\hypersetup{draft}
\begin{document}

\maketitle

\begin{abstract}
We consider the problem of imitation learning from expert demonstrations in partially observable Markov decision processes (POMDPs). Belief representations, which characterize the distribution over the latent states in a POMDP, have been modeled using recurrent neural networks and probabilistic latent variable models, and shown to be effective for reinforcement learning in POMDPs. In this work, we investigate the belief representation learning problem for generative adversarial imitation learning in POMDPs. Instead of training the belief module and the policy separately as suggested in prior work, we learn the belief module jointly with the policy, using a task-aware imitation loss to ensure that the representation is more aligned with the policy's objective. To improve robustness of representation, we introduce several informative belief regularization techniques, including multi-step prediction of dynamics and action-sequences. Evaluated on various partially observable continuous-control locomotion tasks, our belief-module imitation learning approach (BMIL) substantially outperforms several baselines, including the original GAIL algorithm and the task-agnostic belief learning algorithm. Extensive ablation analysis indicates the effectiveness of task-aware belief learning and belief regularization. Code for the project is available online{\parindent=0pt\footnote{\url{https://github.com/tgangwani/BMIL}. Part of this work was done while Tanmay was an intern at Uber AI Labs.}}. 
\end{abstract}

\section{Introduction}
Recent advances in reinforcement learning (RL) have found successful applications in solving complex problems, including robotics, games, dialogue systems, and recommendation systems, among others. Despite such notable success, the application of RL is still quite limited to problems where the observation-space is rich in information and data generation is inexpensive. On the other hand, the environments in real-world problems, such as autonomous driving and robotics, are typically stochastic, complex and partially observable. To achieve robust and practical performance, RL algorithms should adapt to situations where the agent is being fed noisy and incomplete sensory information. To model these types of environments, partially observable Markov decision processes (POMDPs;~\citet{astrom}) have been proposed and widely studied. In a POMDP, since the current observation alone is insufficient for choosing optimal actions, the agent's history (its past observations and actions) is encoded into a {\em belief state}, which is defined as the distribution (representing the agent's beliefs) over the current latent state. Although belief states can be used to derive optimal policies~\citep{kaelbling1998planning, hauskrecht2000value}, maintaining and updating them requires knowledge of the transition and observation models of the POMDP, and is prohibitively expensive for high-dimensional spaces. To overcome this difficulty, several algorithms have been proposed that perform approximate inference of the belief state representation from raw observations,
using recurrent neural networks~\citep{guo2018neural}, variational autoencoders~\citep{igl2018deep, gregor2018temporal}, and Predictive State Representations~\citep{venkatraman2017predictive}. After the belief model has been learned, a policy optimization algorithm is then applied to the belief representation to optimize a predefined reward signal. 

As an alternative to RL from predefined rewards, imitation learning often provides a fast and efficient way for training an agent to complete tasks. Expert demonstrations are provided to guide a learner agent to mimic the actions of the expert without the need to specify a reward function. A large volume of work has been done over the past decades on imitation learning for fully observable MDPs, including the seminal work on generative adversarial imitation learning (GAIL,~\citet{ho2016generative}), but there has been little focus on applying these ideas to partially observable environments.

In this paper, we study the problem of imitation learning for POMDPs. Specifically, we introduce a new belief representation learning approach for generative adversarial imitation learning in POMDPs. Different from previous approaches, where the belief state representation and the policy are trained in a decoupled manner, we learn the belief module jointly with the policy, using a task-aware imitation loss which helps to align the belief representation with the policy's objective.
To avoid potential belief degeneration, we introduce several informative belief regularization techniques, including auxiliary losses of predicting multi-step past/future observations and action-sequences, which improve the robustness of the belief representation. Evaluated on various partially observable continuous-control locomotion tasks built from MuJoCo, our belief-module imitation learning approach (BMIL) substantially outperforms several baselines, including the original GAIL algorithm and the task-agnostic belief learning algorithm. Extensive ablation analysis indicates the effectiveness of task-aware belief learning and belief regularization.

\section{Background and Notation}
\label{sec:backgr}
{\textbf{Reinforcement Learning.}} We consider the RL setting where the environment is modeled as a 
partially-observable Markov decision process (POMDP). A POMDP is characterized by the tuple ($\mathcal{S}$, $\mathcal{A}$, $\mathcal{O}$, $\mathcal{R}$, $\mathcal{T}$, $\mathcal{U}$, , $p(s_0)$, $\gamma$), where $\mathcal{S}$ is the state-space, $\mathcal{A}$ is the action-space, and $\mathcal{O}$ is the observation-space. The true environment states $s_t \in \mathcal{S}$ are latent or unobserved to the agent. Given an action $a_t \in \mathcal{A}$, the next state is governed by the transition dynamics $s_{t+1} \sim \mathcal{T}(s_{t+1}|s_t,a_t)$, an observation is generated as $o_{t+1} \sim \mathcal{U}(o_{t+1}|s_{t+1})$, and reward is computed as $r_t = \mathcal{R}(r_t | s_t,a_t)$. The RL objective involves maximization of the expected discounted sum of rewards, $\eta(\pi_{\theta}) = \mathbb{E}_{p_0, \mathcal{T}, \pi} \big[ \sum_{t=0}^{\infty} \gamma^t r(s_t,a_t) \big]$, where $\gamma\in(0,1]$ is the discount factor, and $p(s_0)$ is the initial state distribution. The action-value function is $Q^{\pi}(s_t, a_t) = \mathbb{E}_{p_0, \mathcal{T}, \pi} \big[ \sum_{t'=t}^{\infty} \gamma^{t'-t} r(s_{t'},a_{t'}) \big]$. We define the unnormalized $\gamma$-discounted state-visitation distribution for a policy $\pi$ by $\rho_{\pi}(s) = \sum_{t=0}^{\infty} \gamma^{t} P(s_t{=}s|\pi)$, where $P(s_t{=}s|\pi)$ is the probability of being in state $s$ at time $t$, when following policy $\pi$ and starting state $s_0 \sim p_0$. The expected policy return $\eta(\pi_{\theta})$ can then be written as $\mathbb{E}_{\rho_{\pi}(s,a)} [r(s,a)]$, where $\rho_{\pi}(s,a) = \rho_{\pi}(s)\pi(a|s)$ is the state-action visitation distribution (also referred to as the occupancy measure). For any policy $\pi$, there is a one-to-one correspondence between $\pi$ and its occupancy measure~\citep{Puterman}. Using the policy gradient theorem~\citep{sutton2000policy}, the gradient of the RL objective can be obtained as $\nabla_{\theta} \eta(\pi_{\theta}) = \mathbb{E}_{\rho_{\pi}(s,a)} \big[\nabla_{\theta}\log \pi_\theta(a|s)Q^{\pi}(s,a)\big]$.

{\textbf{Imitation Learning.}} Learning in popular RL algorithms (such as policy-gradients and Q-learning) is sensitive to the quality of the reward function. In many practical scenarios, the rewards are either unavailable or extremely sparse, leading to difficulty in temporal credit assignment~\citep{sutton1984temporal}. 
In the absence of explicit environmental rewards, a promising approach is to leverage {\em demonstrations} of the completed task by experts, and learn to imitate their behavior. Behavioral cloning (BC; ~\citet{pomerleau1991efficient}) poses imitation as a supervised-learning problem, and learns a policy by maximizing the likelihood of expert-actions in the states visited by the expert. The policies produced with BC are generally not very robust due to the issue of compounding errors; several approaches have been proposed to remedy this~\citep{ross2011reduction, ross2014reinforcement}. Inverse Reinforcement Learning (IRL) presents a more principled approach to imitation by attempting to recover the cost function under which the expert demonstrations are optimal~\citep{ng2000algorithms, ziebart2008maximum}. Most IRL algorithms, however, are difficult to scale up computationally because they require solving an RL problem in their inner loop. Recently,~\citet{ho2016generative} proposed framing imitation learning as an occupancy-measure matching (or divergence minimization) problem. Their architecture (GAIL) forgoes learning the optimal cost function in order to achieve computational tractability and sample-efficiency (in terms of the number of expert demonstrations needed). In detail, if $\rho_{\pi}(s,a)$ and $\rho_{E}(s,a)$ represent the state-action visitation distributions of the policy and the expert, respectively, then minimizing the Jenson-Shanon divergence $\min_{\pi} D_{JS}[\rho_{\pi}(s,a)\ ||\ \rho_{E}(s,a)]$ helps to recover a policy with a similar trajectory distribution as the expert. GAIL iteratively trains a policy ($\pi_\theta$) and a discriminator ($D_\omega$) to optimize the mini-max objective:
\begin{equation}\label{eq:gail}
\begin{aligned}
  \min_\theta \max_\omega \: &\mathbb{E}_{(s,a) \sim \pi, \mathcal{T}} \big[\log (1 - D_{\omega} (s, a)) \big] \\ &+\mathbb{E}_{(s,a) \sim \mathcal{M}_E} \big[\log D_{\omega} (s, a) \big]
\end{aligned}
\end{equation}
where $D_\omega : \mathcal{S} \times \mathcal{A} \rightarrow (0,1)$, $\mathcal{M}_E$ is the buffer with expert demonstrations, and $\mathcal{T}$ is the transition dynamics. 

\section{Methods}
\subsection{Belief Representation in POMDP}\label{subsec:belief_rep}
In a POMDP, the observations are by definition non-Markovian. A policy $\pi(a_t|o_t)$ that chooses actions based on current observations performs sub-optimally, since $o_t$ does not contain sufficient information about the true state of the world. It is useful to infer a distribution on the true states based on the experiences thus far. This is referred to as the {\em belief state}, and is formally defined as the filtering distribution: $p(s_t|o_{\le t}, a_{<t})$. It combines the memory of past experiences with uncertainty about unobserved aspects of the world. Let $h_t := (o_{\le t}, a_{<t})$ denote the observation-action history, and $b_t := \phi(h_t)$ be a function of $h_t$. If $b_t$ is learned such that it forms the sufficient statistics of the filtering posterior over states, i.e., $p(s_t|o_{\le t}, a_{<t}) \approx p(s_t|b_t)$, then $b_t$ could be used as a surrogate code (or representation) for the belief state, and be used to train agents in POMDPs. Henceforth, with slight abuse of notation, we would refer to $b_t$ as the {\em belief}, although it is a high-dimensional representation rather than an explicit distribution over states. 

An intuitive way to obtain this belief is by combining the observation-action history using aggregator functions such as recurrent or convolution networks. For instance, the intermediate hidden states in a recurrent network could represent $b_t$. In the RL setting with environmental rewards, the representation could be trained by conditioning the policy on it, and back-propagating the RL (e.g. policy gradient) loss. However, the RL signal is generally too weak to learn a rich representation $b_t$ that provides sufficient statistics for the filtering posterior over states.~\citet{moreno2018neural} provide empirical evidence of this claim by training oracle models where representation learning is supervised with privileged information in form of the (unknown) environment states, and comparing them with learning solely using the RL loss. The problem is only exacerbated when the environmental rewards are extremely sparse. In our imitation learning setup, the belief update is incorporated into the mini-max objective for adversarial imitation of expert trajectories, and hence the representation is learned with a potentially stronger signal (Section~\ref{subsec:bm}). Prior work has shown that representations can be improved by using auxiliary losses such as reward-prediction~\citep{jaderberg2016reinforcement}, depth-prediction~\citep{mirowski2016learning}, and prediction of future sensory data~\citep{dosovitskiy2016learning, oh2015action}. Inspired by this, in Section~\ref{sec:breg}, we regularize the representation with various prediction losses.  

Recently,~\citet{ha2018recurrent} proposed an architecture (World-Models) that decouples model-learning from policy-optimization. In the model-learning phase, a variational auto-encoder compresses the raw observations to latent-space vectors, which are then temporally integrated using an RNN, combined with a mixture density network. 
In the policy-optimization phase, a policy conditioned on the RNN hidden-states is learned to maximize the rewards. We follow a similar {\em separation-of-concerns} principle, and divide the architecture into two modules: 1) a {\textbf {policy module}} $\pi_{\theta} (a_t|b_t)$ which learns a distribution over actions, conditioned on the belief; and 2) a {\textbf {belief module}} $B_{\phi}$ which learns a good representation of the belief $b_t := B_{\phi}(h_t)$, from the history of observations and actions, $h_t := (o_{\le t}, a_{<t})$. While the policy module is trained with imitation learning, the belief module can be trained in a task-agnostic manner (like in World-Models), or in a task-aware manner. We describe these approaches in following sections.
\subsection{Policy Module}\label{sec:pol_module}
The goal of our agent is to learn a policy by imitating a few expert demonstration trajectories of the form  $\{o_i, a_i\}_{i=0}^{|\tau|}$. Similar to the objective in GAIL, we hope to minimize the Jenson-Shanon divergence between the state-action visitation distributions of the policy and the expert: $\min_{\pi} D_{JS}[\rho_{\pi}(s,a)\ ||\ \rho_{E}(s,a)]$. However, since the true environment state $s_t$ is unobserved in POMDPs, we modify the objective to involve the belief representation $b_t$ instead, since it characterizes the posterior over $s_t$ via the generative process $p(s_t|b_t)$. Defining the {\em belief-visitation} distribution $\rho_{\pi}(b)$ for a policy analogously to the state-visitation distribution, the data processing inequality for $f$-divergences provides that: $D_{JS} [\rho_{\pi}(s)\ ||\ \rho_{E}(s)] \le D_{JS} [\rho_{\pi}(b)\ ||\ \rho_{E}(b)]$.
%
%
Please see Appendix~\ref{appn:proofs} for the proof. The objective $\min_{\pi} D_{JS} [\rho_{\pi}(b)\ ||\ \rho_{E}(b)]$ thus minimizes an upper bound on the $D_{JS}$ between the state-visitation distributions of the expert and the policy. A further relaxation of this objective allows us to explicitly include the belief-conditioned policy $\pi(a|b)$ into the divergence minimization objective: $D_{JS} [\rho_{\pi}(b)\ ||\ \rho_{E}(b)] \le D_{JS} [\rho_{\pi}(b, a)\ ||\ \rho_{E}(b,a)]$, 
%
%
where $\rho_{\pi}(b, a) = \rho_{\pi}(b) \pi(a|b)$ is the belief-action visitation (proof in  Appendix~\ref{appn:proofs}).

\begin{figure}[t]
    \begin{center}
    \includegraphics[width=.4\textwidth]{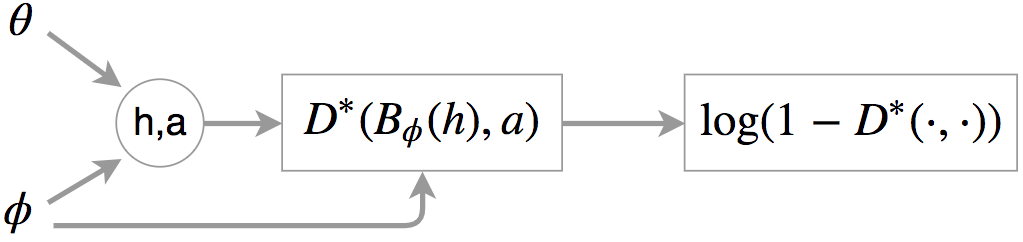}
    \caption{\small Stochastic computation graph for the expectation: $\widetilde{\mathbb{E}}_{(b,a) \sim \pi, \mathcal{T}} \big[\log (1 - D^{*} (b, a)) \big]$. Both the policy $(\theta)$ and belief module $(\phi)$ parameters influence the generation of observation-action sequences, $(h_t ,a_t) := \{o_{\le t}, a_{\le t}\}$ through environment interaction. $b = B_\phi(h)$ is the belief. Circles represent stochastic nodes; rectangles are deterministic nodes.}
    \label{fig:scg}
    \end{center}
\end{figure}

\textbf{Minimizing $\mathbf{D_{JS} [\rho_{\pi}(b, a)\ ||\ \rho_{E}(b,a)]}$}. Although explicitly formulating these visitation distributions is difficult, it is possible to obtain an empirical distribution of $\rho_\pi(b,a)$ by rolling out trajectories $(o_1, a_1, \dotsc)$ from $\pi$, and using our belief module to produce samples of belief-action tuples $(b_t, a_t)$, where $b_t := B_{\phi}(h_t), h_t := (o_{\le t}, a_{<t})$. Similarly, the expert demonstrations buffer $\mathcal{M}_E$ contains observation-actions sequences, and can be used as an estimate of $\rho_E(b,a)$. Therefore, $D_{JS}$~\footnote{To reduce clutter, we shorthand $D_{JS} [\rho_{\pi}(b, a)\ ||\ \rho_{E}(b,a)]$ with just $D_{JS}$ for the remainder of this section.} can be approximated (up to a constant shift) with a binary classification problem as exploited in GANs~\citep{goodfellow2014generative}:
\begin{equation}\label{eq:djs}
\begin{aligned}
  D_{JS}(\theta;\phi) &\approx \max_{\omega}
  \widetilde{\mathbb{E}}_{(b,a) \sim \mathcal{M}_E} \big[\log D_{\omega} (b, a) \big]
  \\ &+  \widetilde{\mathbb{E}}_{(b,a) \sim \pi, \mathcal{T}} \big[\log (1 - D_{\omega} (b, a)) \big]
\end{aligned}
\end{equation}
where $\theta$ are the parameters for the policy $\pi_\theta(a|b)$, $D_\omega$ is the discriminator, and $\mathcal{T}$ is the transition dynamics. It should be noted that $D_{JS}$ is a function of the belief module parameters $\phi$ through its dependence on the belief states. The imitation learning objective for optimizing the policy is then obtained as:
\begin{equation}\label{eq:pol_opt}
\begin{aligned}
  \min_{\theta} D_{JS}(\theta;\phi) &\approx \min_{\theta} \max_{\omega}
  \widetilde{\mathbb{E}}_{(b,a) \sim \mathcal{M}_E} \big[\log D_{\omega} (b, a) \big]
  \\ &+  \widetilde{\mathbb{E}}_{(b,a) \sim \pi, \mathcal{T}} \big[\log (1 - D_{\omega} (b, a)) \big]
\end{aligned}
\end{equation}
 In Equation~\ref{eq:djs}, denoting the functional maximum over $D_\omega$ by $D^{*}$, the gradient for policy optimization is: $\nabla_\theta D_{JS}(\theta;\phi) \approx \nabla_{\theta} \widetilde{\mathbb{E}}_{(b,a) \sim \pi, \mathcal{T}} \big[\log (1 - D^{*} (b, a)) \big]$. Figure~\ref{fig:scg} shows the stochastic computation graph~\citep{schulman2015gradient} for this expectation term, where the stochastic nodes are represented by circles, deterministic nodes by rectangles, and we have written belief as a function of the history. Given fixed belief module parameters ($\phi$), the required gradient w.r.t $\theta$ is obtained using the policy gradient theorem~\citep{sutton2000policy} as:
\begin{equation}\label{eq:policy_gradient}
\begin{aligned}
    & \nabla_\theta D_{JS}(\theta;\phi) \approx \nabla_{\theta} \widetilde{\mathbb{E}}_{(b,a) \sim \pi, \mathcal{T}} \big[\log (1 - D^{*} (b, a)) \big] \\
    & \qquad \qquad = \widetilde{\mathbb{E}}_{(b,a) \sim \pi, \mathcal{T}} \big[\nabla_{\theta}\log \pi_{\theta}(a|b)\hat{Q}^{\pi}(b,a)\big], \text{where} \\
    & \hat{Q}^{\pi}(b_t,a_t) = \widetilde{\mathbb{E}}_{(b,a) \sim \pi, \mathcal{T}} \big[ \sum_{t'=t}^{\infty} \gamma^{t'-t} \log (1 - D^{*} (b_{t'}, a_{t'})) \big] 
\end{aligned}
\end{equation}
Therefore, updating the policy to minimize $D_{JS}$ is approximately the same as applying the standard policy-gradient using the rewards obtained from a learned discriminator, $r(b,a) = -\log(1-D^{*}(b,a))$. As is standard practice, we do not train the discriminator to optimality, but 
rather jointly train the policy and discriminator using iterative gradient updates. The discriminator is updated using the gradient from Equation~\ref{eq:djs}, while the policy is updated with gradient from Equation~\ref{eq:policy_gradient}. We now detail the update rule for $\phi$. 

\subsection{Belief Module}\label{subsec:bm}
This module transforms the history ($o_{\le t}, a_{<t}$) into a belief representation. Various approaches could be used to aggregate historical context, such as RNNs, masked convolutions~\citep{gehring2017convolutional} and attention-based methods~\citep{vaswani2017attention}. In our implementation, we model the belief module $B_\phi$ with an RNN, such that $b_t = B_\phi (b_{t-1}, o_t, a_{t-1})$. We use GRUs~\citep{cho2014properties} as they have been demonstrated to have good empirical performance. We denote by $\mathcal{R}$, a replay-buffer which stores observation-action sequences (current and past) from the agent. As stated before, the belief module could be learnt in a task-agnostic manner (similar to~\citet{ha2018recurrent}), or with task-awareness.

\textbf{Task-agnostic learning (separately from policy).} An unsupervised approach to learning $\phi$ without accounting for the agent's objective, is to maximize the joint likelihood of the observation sequence, conditioned on the actions, $\log p(o_{\le T}|a_{<T})$. This decomposes autoregressively as $\sum_t \log p(o_t|o_{<t},a_{<t})$. The objective can be optimized by conditioning a generative model for $o_t$ on the RNN hidden state $b^{\phi}_{t-1}$ and action $a_{t-1}$, and using MLE. Using a unimodal Gaussian  generative model (learned function $g$ for the mean, and fixed variance), the autoregressive loss to minimize is:
\begin{equation}
    \mathcal{L}^{AR}(\phi) = \mathbb{E}_{\mathcal{R}} ||o_t - g(b^{\phi}_{t-1}, a_{t-1})||^2_2 
\end{equation}
%
%
\textbf{Task-aware learning (jointly with policy).}
Since the policy is conditioned on the belief, an intuitive way to improve the agent's performance is to learn the belief with an objective more aligned with policy-learning. Since the agent minimizes $D_{JS}(\theta,\phi)$, as defined in Equation~\ref{eq:djs}, the same imitation learning objective naturally can also be used for learning $\phi$:  
\begin{equation}\label{eq:loss_im}
\begin{aligned}
  \mathcal{L}^{IM}(&\phi) := D_{JS}(\theta,\phi) \approx \\
  &\widetilde{\mathbb{E}}_{(h,a) \sim \mathcal{M}_E} \big[\log D^{*} (B_{\phi}(h), a) \big] \\
  &+\widetilde{\mathbb{E}}_{(h,a) \sim \pi_{\theta}(a|B_{\phi}(h)), \mathcal{T}} \big[\log (1 - D^{*} (B_{\phi}(h), a)) \big] 
\end{aligned}
\end{equation}
which is the same as Equation~\ref{eq:djs} except for the use of the optimal discriminator ($D^{*}$), and that we have written the belief in terms of history $b := B_{\phi}(h)$ to explicitly bring out the dependence on $\phi$. The gradient of the first expectation term w.r.t $\phi$ is straightforward; the gradient of the second expectation term w.r.t $\phi$ (for given fixed parameters $\theta$) comprises of a policy-gradient term and a pathwise-derivative term (Figure~\ref{fig:scg}). Therefore, $\nabla_\phi D_{JS}(\theta;\phi)$ can be approximated with:
\begin{equation}\label{eq:grad_wrt_phi}
\begin{aligned}
    &\widetilde{\mathbb{E}}_{(h,a) \sim \mathcal{M}_E} \big[\nabla_{\phi} \log D^{*} (B_{\phi}(h), a) \big] \\ &+ \:
    \underbrace{\widetilde{\mathbb{E}}_{(h,a) \sim \pi_{\theta}(a|B_{\phi}(h)), \mathcal{T}} 
    \big[\nabla_{\phi}\log \pi_{\theta}(a|B_{\phi}(h))\hat{Q}^{\pi}\big]}_{\text{policy-gradient term}}\\ &+ \: \underbrace{\widetilde{\mathbb{E}}_{(h,a) \sim \pi_{\theta}(a|B_{\phi}(h)), \mathcal{T}} \big[\nabla_{\phi} \log (1 - D^{*} (B_{\phi}(h), a)) \big]}_{\text{pathwise-derivate term}} 
\end{aligned}
\end{equation}
where $\hat{Q}$ is as defined in Equation~\ref{eq:policy_gradient}. The overall mini-max objective for jointly training the policy, belief and discriminator is:
\begin{equation}\label{eq:overall_minimax}
\begin{aligned}
  \min_{\phi, \theta} \max_{\omega}\:
  &\widetilde{\mathbb{E}}_{(b,a) \sim \mathcal{M}_E} \big[\log D_{\omega} (b, a) \big] \\
  &+ \: \widetilde{\mathbb{E}}_{(b,a) \sim \pi, \mathcal{T}} \big[\log (1 - D_{\omega} (b, a)) \big] \\
\end{aligned}
\end{equation}
\begin{figure*}[t]
    \begin{center}
    \includegraphics[width=\textwidth]{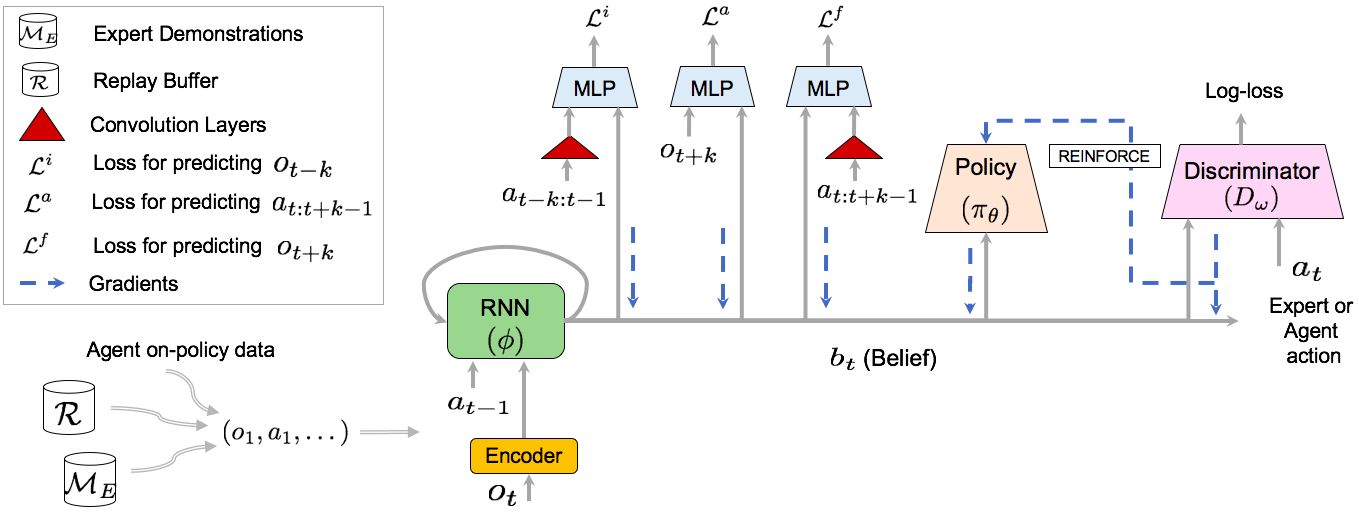}
    \caption{Schematic diagram of our complete architecture. The belief module $B_\phi$ is a recurrent network with GRU cells, and encodes trajectories $(o_1, a_1, \dotsc)$ from agent (on-policy data), replay buffer $\mathcal{R}$ (off-policy data) and expert demonstrations memory $\mathcal{M}_E$ into belief representations ($b_t$). $B_\phi$ is updated with imitation-loss (Equation~\ref{eq:loss_im}) computed from the current policy and discriminator networks. It is further regularized with forward-, inverse- and action-regularization using MLPs (colored in blue in the figure). Convolution layers (colored in red) encode the past actions $(a_{t-k:t-1})$
    and future actions $(a_{t:t+k-1})$ into compact representations, which are then fed into the MLPs. The policy $\pi_\theta(a_t|b_t^{\phi})$ is conditioned on the belief, and updated using imitation learning (Equation~\ref{eq:policy_gradient}). The discriminator $D_{\omega}(b_t^{\phi},a_t)$ is a binary classifier trained on tuples from the agent and expert demonstrations (Equation~\ref{eq:djs}).}
    \label{fig:full_schema}
    \end{center}
\end{figure*}

\subsection{Belief Regularization}\label{sec:breg}
With the mini-max objective (Equation~\ref{eq:overall_minimax}), it may be possible that the belief parameters ($\phi$) are driven towards a degenerate solution that ignores the history $(o_{\le t}, a_{<t})$, thereby producing constant (or similar) beliefs for policy and expert trajectories. Indeed, if we omit the actions ($a$) in the discriminator $D_\omega(b,a)$, a constant belief output is an optimal solution for Equation~\ref{eq:overall_minimax}. To learn a belief representation that captures relevant historical context and is useful for deriving optimal policies, we add forward-, inverse- and action-regularization to the belief module. We define and motivate them from the perspective of mutual information maximization.

\textbf{Notation.}
For two continuous random variables $X, Y$, mutual information is defined as $I(X;Y) := H(X) - H(X|Y)$, where $H$ denotes the differential entropy~\footnote{The usual notation for differential entropy ($h$) is not used to avoid confusion with the history $h_t$ used in previous sections.}. Intuitively, $I(X;Y)$ measures the dependence between $X$ and $Y$. Conditional mutual information is defined as $I(X;Y|Z) := \mathbb{E}_z[(I(X;Y)|z]$. Given $Y$, if $X$ and $Z$ are independent ($X \perp Z | Y$), then $X,Y,Z$ form a Markov Chain ($X \shortrightarrow Y \shortrightarrow Z$), and the data processing inequality for mutual information states that $I(X;Z) \le I(X;Y)$.

\textbf{Forward regularization.}  
As discussed in Section~\ref{subsec:belief_rep}, an ideal belief representation completely characterizes the posterior over the true environment states $p(s_t|b_t)$. Therefore, it ought to be correlated with future true states ($s_{t+k}$), conditioned on the intervening future actions $(a_{t:t+k-1})$. We frame this objective as maximization of the following conditional mutual information: $I(b_t;s_{t+k}|a_{t:t+k-1})$. Since $o_{t+k} \perp b_t | s_{t+k}$ because of the observation generation process in a POMDP, we get the following after using the data processing inequality for mutual information:
\begin{equation}\label{eq:mi_1}
     \begin{aligned}
    I(b_t;&s_{t+k}|a_{t:t+k-1}) \ge I(b_t;o_{t+k}|a_{t:t+k-1}) \\
    &=  \mathbb{E}_{a_{t:t+k-1}} \big[ H(o_{t+k} | a_{t:t+k-1}) \\ &\qquad - H(o_{t+k} | b_t; a_{t:t+k-1}) \big] \\
    &\ge \mathbb{E}_{a_{t:t+k-1}} \bigg[ H(o_{t+k} | a_{t:t+k-1}) \\ &\qquad + \mathbb{E}_{o_{t+k}, b_t} \big[ \log q (o_{t+k} | b_t; a_{t:t+k-1}) \big] \bigg]
     \end{aligned}
\end{equation}
where the final inequality follows because we can lower bound the mutual information using a variational approximation $q$, similar to the variational information maximization algorithm~\citep{agakov2004variational}.
%
%
Therefore, we maximize a lower bound to the mutual information $I(b_t;s_{t+k}|a_{t:t+k-1})$ with the surrogate objective:
\begin{equation*}
  \max_{\phi, q}  \mathbb{E}_{\substack{o_{t+k}, b_t, \\ a_{t:t+k-1}}} \big[ \log q (o_{t+k} | b^{\phi}_t; a_{t:t+k-1}) \big]
\end{equation*}
With the choice of a unimodal Gaussian (learned function $g$ for the mean, and fixed variance) for the variational distribution $q$, the loss function for forward regularization of the belief module is:
\begin{equation*}
    \mathcal{L}^{f}(\phi) = \mathbb{E}_{\mathcal{R}} ||o_{t+k} - g(b^{\phi}_{t}, a_{t:t+k-1})||^2_2 
\end{equation*}
where the expectation is over trajectories $(o_1, a_1, \dotsc)$ sampled from the replay buffer $\mathcal{R}$.

\textbf{Inverse regularization.} It is desirable that the belief at time $t$ is correlated with the past true states ($s_{t-k}$), conditioned on the intervening past actions $(a_{t-k:t-1})$. This should improve the belief representation by helping to capture long-range dependencies. Proceeding in a manner similar to above, the conditional mutual information between these signals, $I(s_{t-k};b_{t}|a_{t-k:t-1})$, can be lower bounded by $I(o_{t-k};b_{t}|a_{t-k:t-1})$ using the data processing inequality. Again, as before, this can be further lower bounded using a variational distribution $q$ for generating past observation $o_{t-k}$. A unimodal Gaussian (mean function $g$) for $q$ yields the following loss  for inverse regularization, which is optimized using trajectories from the replay $\mathcal{R}$:
\begin{equation*}
    \mathcal{L}^{i}(\phi) = \mathbb{E}_{\mathcal{R}} ||o_{t-k} - g(b^{\phi}_{t}, a_{t-k:t-1})||^2_2 
\end{equation*}
\textbf{Action regularization.} We also wish to maximize $I(a_{t:t+k-1}; s_{t+k}\ |\ b_t)$ for the reason that, conditioned on the current belief $b_t$, a sequence of $k$ subsequent actions $(a_{t:t+k-1})$ should provide information about the resulting true future state ($s_{t+k}$). Similar lower bounding and use of a variational distribution with mean function $g$ for generating action-sequences gives us the loss:
\begin{equation*}
    \mathcal{L}^{a}(\phi) = \mathbb{E}_{\mathcal{R}} ||(a_{t:t+k-1}) - g(b^{\phi}_{t}, o_{t+k})||^2_2 
\end{equation*}
The complete loss function for training the belief module results from a weighted combination of the imitation-loss and regularization terms. Imitation-loss uses on-policy data and expert demonstrations $(\mathcal{M}_E)$, while the regularization losses are computed with on-policy and off-policy data, as well as $\mathcal{M}_E$.
\begin{equation}\label{eq:belief_losses}
    \mathcal{L}(\phi) = \mathcal{L}^{IM} + \lambda_1\mathcal{L}^{f}+ \lambda_2\mathcal{L}^{i} + \lambda_3\mathcal{L}^{a}
\end{equation}
We derive our final expressions for $(\mathcal{L}^{f}, \mathcal{L}^{i}, \mathcal{L}^{a})$ by modeling the respective variational distributions $(q)$ as fixed-variance, unimodal Gaussians. We later show that using this simple model results in appreciable performance benefits for imitation learning. Other expressive model classes, such as mixture density networks and flow-based models~\citep{rezende2015variational}, can be readily used as well, to learn complex and multi-modal distributions over the future observations $(o_{t+k})$, past observations $(o_{t-k})$ and action-sequences $(a_{t:t+k-1})$.
\subsection{Learning Algorithm}
Figure~\ref{fig:full_schema} shows the schematic diagram of our complete architecture, including an overview of implemented neural networks. In Algorithm~\ref{algo:simple}, we outline the major steps of the training procedure. In each iteration, we run the policy  
for a few steps and obtain shaped rewards from the current discriminator (Line 6). The policy parameters are then updated using A2C, which is the synchronous adaptation of asynchronous advantage actor-critic (A3C;~\citet{mnih2016asynchronous}), as the policy-gradient algorithm (Line 10). Other RL algorithms, such as those based on trust-regions methods~\citep{schulman2015trust} could also be readily used. Similar to the policy (actor), the baseline (critic) used for reducing variance of the stochastic gradient-estimation is also conditioned on the belief. To further reduce variance, Generalized Advantage Estimation (GAE;~\citet{schulman2015high}) is used to compute the advantage. Apart from the policy-gradient, on-policy data also enables computing the gradient for the discriminator network (Line 13) and the belief module (Line 14). The belief is further refined by minimizing the regularization losses on off-policy data from the replay buffer $\mathcal{R}$ (Line 15).

The regularization losses $(\mathcal{L}^{f}, \mathcal{L}^{i}, \mathcal{L}^{a})$ described in Section~\ref{sec:breg} include a hyperparameter $k$ that controls the temporal offset of the predictions. For instance, for $\mathcal{L}^{i}(\phi;k)$, the larger the $k$, the farther back in time the observation predictions are made, conditioned on the current belief and past actions. The temporal abstractions provided by multi-step predictions $(k{>}1)$ should help to extract more global information from the input stream into the belief representation. Our ablations (Section~\ref{sec:exp}) show the performance benefit of including multi-step losses. Various strategies for selecting $k$ are possible, such as uniform sampling from a range~\citep{guo2018neural} and adaptive selection based on a curriculum~\citep{oh2015action}. For simplicity, we choose fixed values, and leave the exploration of the more sophisticated approaches to future work. Hence, our total regularization loss comprises of single-step $(k{=}1)$ and multi-step $(k{=}5)$ forward-, inverse-, and action-prediction losses. For encoding a sequence of past or future actions into a compact representation, we use multi-layer convolution networks (Figure~\ref{fig:full_schema}).

\newcommand{\xCommentSty}[1]{\ttfamily\textcolor{blue}{#1}}
\SetCommentSty{xCommentSty}

\RestyleAlgo{ruled}
\LinesNumbered
\begin{algorithm}[h]
\SetNoFillComment

 \BlankLine
 \For{each iteration}{
    \BlankLine
    $d_{\pi}$ = \{\}, $d_{E}$ = \{\} \\
    
    \BlankLine
    \tcc{Rollout $c$ steps from policy}
    \Repeat{$|d_{\pi}| == c$}{
        Get observation $o_t$ from environment\\
        $a_t \sim \pi_{\theta}(a_t|b_t)$, where $b_t = B_{\phi}(o_{\le t}, a_{<t})$ \\
        $r_t = - \log ( 1 - D_{\omega}(b_t, a_t))$ \\
        $d_{\pi} \leftarrow d_{\pi} \cup (b_t, a_t, r_t)$ \\
        If $o_t$ is terminal, add rollout $\{o_i, a_i\}_{i=0}^{|\tau|}$ to $\mathcal{R}$
        }
    
    \BlankLine
    \tcc{Update Policy}
    Update $\theta$ with policy-gradient (Eq.~\ref{eq:policy_gradient})
    
    \BlankLine
    \BlankLine
    \tcc{Update discriminator $\omega$}
    Fetch ($o_t, a_t, \dotsc$) of length $c$ from $\mathcal{M}_E$ \\
    Generate belief-action tuples $d_{E} = \{(b_i, a_i)\}_{i=t}^{t+c-1}$ \\
    Update $\omega$ with log-loss objective using $d_{\pi}$ and $d_{E}$

    \BlankLine
    \BlankLine
    \tcc{Update Belief Module $\phi$}
    Update $\phi$ with $\nabla_{\phi} \mathcal{L}(\phi)$ using $d_{\pi}$ and $d_{E}$ (Eq.~\ref{eq:belief_losses}) 
     
    \BlankLine
    \BlankLine
    \tcc{Off-policy Updates}
    \For{few update steps}{
    Fetch ($o_t, a_t, \dotsc$) of length $c$ from $\mathcal{R}$  \\
    Update $\phi$ with $\nabla_{\phi} (\lambda_1\mathcal{L}^{f}+ \lambda_2\mathcal{L}^{i} + \lambda_3\mathcal{L}^{a})$
    }
 }
 \caption{Belief-module Imitation Learning (BMIL)}
 \label{algo:simple}
\end{algorithm}

\vspace{-2mm}
\section{Related Work}
\label{sec:rel}
While we cannot do full justice to the extensive literature on algorithms for training agents in POMDPs, we here mention some recent related work. Most prior algorithms for POMDPs assume access to a predefined reward function. These include approaches based on Q-learning (DRQN;~\citet{hausknecht2015deep}), policy-gradients~\citep{igl2018deep}, partially observed guided policy search~\citep{zhang2016learning}, and planning methods~\citep{silver2010monte,ross2008online,pineau2003point}. In contrast, we propose to adapt ideas from generative adversarial imitation learning to learn policies in POMDPs without environmental rewards.

Learning belief states from history $(o_{\le t}, a_{<t})$ was recently explored in~\citet{guo2018neural}. The authors show that training the belief representation with a Contrastive Predictive Coding (CPC,~\citet{oord2018representation}) loss on future observations, conditioned on future actions, helps to infer knowledge about the underlying state of the environment. Predictive State Representations (PSRs) offer another approach to modeling the belief state in terms of observable data~\citep{littman2002predictive}. The assumption in PSRs is that the filtering distribution can be approximated with a distribution over the $k$ future observations, conditioned on future actions, $p(s_t|o_{\le t}, a_{<t}) \approx p(o_{t+1:t+k}|o_{\le t}, a_{<{t+k}})$. PSRs combined with RNNs have been shown to improve representations by predicting future observations~\citep{venkatraman2017predictive, hefny2018recurrent}. 
While we also make future predictions, a key difference compared to aforementioned methods is that our belief representation is additionally regularized by predictions in the past, and in action-space, which we later show benefits our approach.

State-space models (SSMs;~\citet{fraccaro2016sequential, goyal2017z, buesing2018learning}), which represent the unobserved environment states with latent variables, have also been used to obtain belief states.
~\citet{igl2018deep} use a particle-filtering method to train a VAE, and represent the belief state with a weighted collection of particles. The model is also updated with the RL-loss using a belief-conditioned policy.
~\citet{gregor2018temporal} proposed TD-VAE, which explicitly connects belief distributions at two distant timesteps, and enforces consistency between them using a transition distribution and smoothing posterior. Although we use a deterministic model for our belief module $(B_\phi)$, our methods apply straightforwardly to SSMs as well.

\section{Experiments}
\label{sec:exp}
\begin{figure*}[t]
    \begin{center}
    \includegraphics[width=\textwidth]{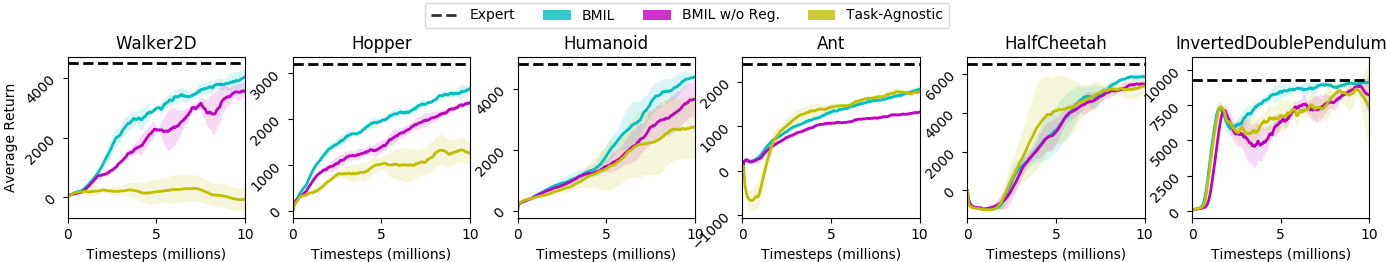}
    \caption{\small Mean episode-returns vs. timesteps of environment interaction. BMIL is our proposed architecture (Figure~\ref{fig:full_schema}); BMIL w/o Reg excludes the various regularization terms (Section~\ref{sec:breg}) from this design; {\em Task-Agnostic} learns the belief module separately from the policy using a task-agnostic loss ($\mathcal{L}^{AR}$, Section~\ref{subsec:bm}). We plot the average and standard-deviation over 5 random seeds.}
    \label{fig:perf}
    \end{center}
\end{figure*}

The goal in this section is to evaluate and analyze the performance of our proposed architecture for imitation learning in partially-observable environments, given some expert demonstrations. Herein, we describe our environments, provide comparisons with GAIL, and perform ablations to study the importance of the design decisions that motivate our architecture.

\begin{figure}[h]
    \centering
    \includegraphics[width=0.4\textwidth]{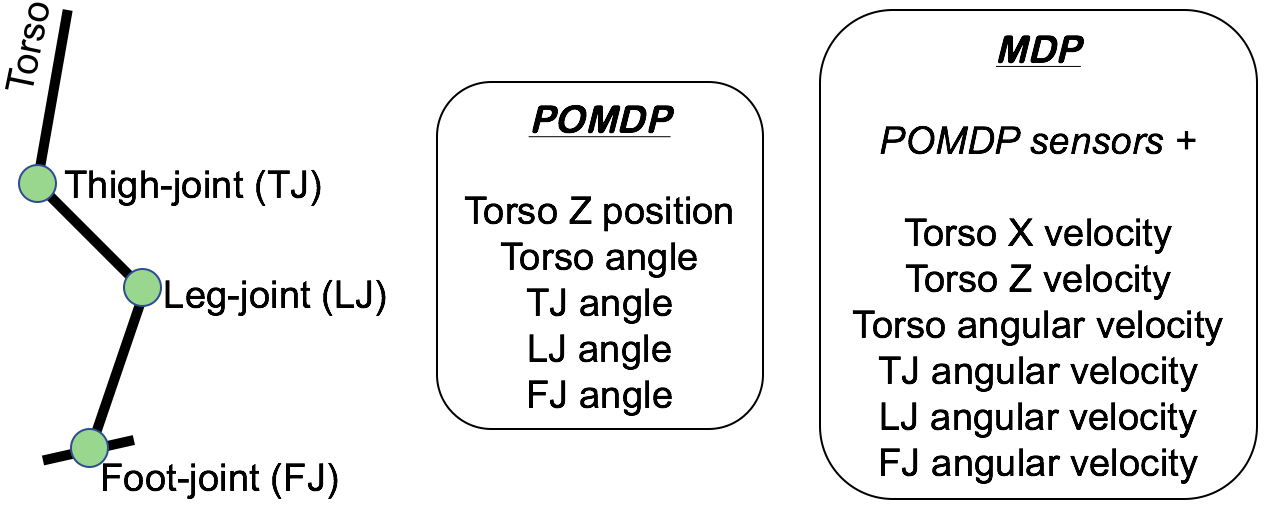}
    \caption{\small{Comparison of sensor information available to the agent in the MDP (original) and the POMDP (modified) settings for Hopper-v2 from the Gym MuJoCo suite.}}
    \label{fig:hopper_schema}
\end{figure}

{\textbf{Partially-observable locomotion tasks.}} We benchmark high-dimensional, continuous-control locomotion environments based on the MuJoCo physics simulator, available in OpenAI Gym~\citep{brockman2016openai}. 
%
%
The standard Gym MuJoCo library of tasks, however, consists of MDPs (and not POMDPs), since observations in such tasks contain sufficient state-information to learn an optimal policy conditioned on only the current observation. As such, it has been extensively used to evaluate performance of reinforcement-learning and imitation-learning algorithms in the MDP setting~\citep{schulman2017proximal, ho2016generative}. To transform these tasks into POMDPs, we follow an approach similar to~\citet{duan2016benchmarking}, and redact some sensory data from the observations. Specifically, from the default observations, we remove measurements for the translation and angular velocities of the torso, and also the velocities for all the link joints. We denote the original (MDP) observations by $s \in \mathcal{S}$, and the curtailed (POMDP) observations by $o \in \mathcal{O}$. Figure~\ref{fig:hopper_schema} shows the {\em Hopper} agent composed of 4 links connected via actuated joints, and describes the original MDP sensors and our POMDP modifications. Similar information for other MuJoCo tasks is included in Appendix~\ref{appn:sensor_info}.

For all experiments, we assume access to 50 expert demonstrations of the type $\{o_i, a_i\}_{i=0}^{|\tau|}$, for each of the tasks. The policy and discriminator networks are feed-forward MLPs with two 64-unit layers. 
The policy network outputs include the action mean and per-action variances (i.e.\ actions are assumed to have an independent Gaussian distribution). In the belief module, the dimension of the GRU cell is 256, while the MLPs used for regularization have two 64-unit feed-forward layers. More details and hyperparameters are in Appendix~\ref{appn:hyperparams}.

\subsection{Comparison to GAIL}
Our first baseline is modeled after the architecture used in the original GAIL approach~\citep{ho2016generative}. It consists of feed-forward policy and discriminator networks, without the recurrent belief module. The policy is conditioned on $o_t$, and the discriminator performs binary classification on $(o_t, a_t$) tuples. The update rules for the policy and discriminator are obtained in similar way as Equation~\ref{eq:policy_gradient} and Equation~\ref{eq:djs}, respectively, by replacing the belief $b_t$ with observation $o_t$. The next baseline, referred to as GAIL+Obs. stack, augments GAIL by concatenating 3 previous observations to each $o_t$, and feeding the entire stack as input to the policy and discriminator networks. This approach has been found to extract useful historical context for a better state-representation~\citep{mnih2015human}. We abbreviate our complete proposed architecture (Figure~\ref{fig:full_schema}) by BMIL, short for Belief-Module Imitation Learning. BMIL jointly trains the policy, belief and discriminator networks using a mini-max objective (Equation~\ref{eq:overall_minimax}), and additionally regularizes the belief with multi-step predictions. Table~\ref{table:gail_comparison} compares the performance of different designs on POMDP MuJoCo. We shows the mean episode-returns, averaged over 5 runs with random seeds, after 10M timesteps of interaction with the environment. We observe that GAIL---both with and without observation stacking---is unable to successfully imitate the expert behavior. Since the observation $o_t$ alone does not contain adequate information, the policy conditioned on it performs sub-optimally. Also, the discriminator trained on $(o_t, a_t$) tuples does not provide robust shaped rewards. Using the full state $s_t$ instead of $o_t$ in our experiments leads to successful imitation with GAIL, suggesting that the performance drop in Table~\ref{table:gail_comparison} is due to partial observability, rather than other artifacts such as insufficient network capacity, or lack of algorithmic or hyperparameter tuning. Further, we see that techniques such as stacking past observations provide only a marginal improvement in some of the tasks. In contrast, in BMIL, the belief module curates a belief representation from the history ($o_{\le t}, a_{<t}$), which is used both for discriminator training, and to condition the action-distribution (policy). BMIL achieves scores very close to those of the expert.

\begin{table}[h]
\centering
\setlength\belowcaptionskip{-15pt}
\resizebox{\linewidth}{!}{%
\begin{tabular}{c||c|c|c|c}
                       & GAIL & {\vtop{\hbox{\strut \;\;\;\; GAIL + }\hbox{\strut Obs. stack}}} &  {\vtop{\hbox{\strut BMIL}\hbox{\strut (Ours)}}} & {\vtop{\hbox{\strut \; Expert}\hbox{\strut ($\approx$ Avg.)}}} \\ \hline \hline
                       
Inv.DoublePend.  & 109 & 1351 & \textbf{9104} &  9300     \\  
Hopper      & 157 & 517 & \textbf{2665} &  3200     \\ 
Ant & 895 & 1056 & \textbf{1832} & 2400    \\
Walker & 357 & 562 & \textbf{4038} &  4500      \\           
Humanoid      & 1686 &  1284 &  \textbf{4382} & 4800        \\   
Half-cheetah   & 205 & -948 & \textbf{5860} &  6500       
\end{tabular}}
\caption{\small Mean episode-returns, averaged over 5 runs with random seeds, after 10M timesteps in POMDP MuJoCo.}
\label{table:gail_comparison}
\end{table}

\subsection{Comparison to GAIL with Recurrent Networks}
For our next two baselines, we replace the feed-forward networks in GAIL with GRUs. GAIL-RF uses a recurrent policy and a feed-forward discriminator, while in GAIL-RR, both the policy and the discriminator are recurrent. In both these baselines, the belief is created internally in the recurrent policy module. Importantly, unlike BMIL, the belief is not shared between the policy and the discriminator. The average final performance of GAIL-RF and GAIL-RR in our POMDP environments is shown in Table~\ref{table:gail_recur_comparison}. We observe that GAIL-RR does not perform well on most of the tasks. A plausible explanation for this is that using the adversarial binary classification loss for training the discriminator parameters does not induce a sufficient representation of the belief state in its recurrent network. The other baseline, GAIL-RF, avoids this issue with a feed-forward discriminator trained on $(o_t, a_t$) tuples from the expert and the policy. This leads to much better performance. However, BMIL consistently outperforms GAIL-RF, most significantly in {\em Humanoid} (1.6$\times$ higher score), indicating the effectiveness of the decoupled architecture and other design decisions that motivate BMIL. Figure~\ref{fig:perf_gail_recurr} (in Appendix) plots the learning curves for these experiments. Appendix~\ref{appn:difficult_exp} further shows that BMIL outperforms the strongest baseline (GAIL-RF) on additional environments with accentuated partial observability.

\begin{table}[h]
\centering
\setlength\belowcaptionskip{-15pt}
\resizebox{0.85\linewidth}{!}{%
\begin{tabular}{c||c|c|c}
                       & GAIL-RR & GAIL-RF &  {\vtop{\hbox{\strut BMIL}\hbox{\strut (Ours)}}} \\ \hline \hline
                       
Inv.DoublePend.  & 8965 & 9103 & \textbf{9104}       \\  
Hopper      & 955 & 2164 & \textbf{2665}       \\ 
Ant & -533 & 1612 & \textbf{1832}    \\
Walker & 400 & 3188 & \textbf{4038}        \\           
Humanoid      & 3829 &  2761 &  \textbf{4382}       \\   
Half-cheetah   & -922 & 5011 & \textbf{5860}        
\end{tabular}}
\caption{\small Mean episode-returns, averaged over 5 runs with random seeds, after 10M timesteps in POMDP MuJoCo.}
\label{table:gail_recur_comparison}
\end{table}

\subsection{Ablation Studies}

\textbf{How crucial is belief regularization?} To quantify this, we compare the performance of our architecture with and without belief regularization (BMIL vs. BMIL w/o Reg.). Figure~\ref{fig:perf} plots the mean episode-returns vs. timesteps of environment interaction, over the course of learning. We observe that including regularization leads to better episode-returns and sample-complexity for most of the tasks considered, indicating that it is useful for shaping the belief representations. The single- and multi-step predictions for $(\mathcal{L}^{f}, \mathcal{L}^{i})$ are made in the observation space. Although we see good improvements for tasks with high-dimensional spaces, such as Humanoid $(|\mathcal{O}|{=}269)$ and Ant $(|\mathcal{O}|{=}97)$, predicting an entire raw observation may be sub-optimal and computationally wasteful for some tasks, since it requires modeling the complex structures within an observation. To avoid this, predictions can be made in a compact encoding space (output of Encoder in Figure~\ref{fig:full_schema}). In Appendix~\ref{appn:encoding_pred}, we show the performance of BMIL with predictions in encoding-space rather than observation-space, and note that the scores are quite similar.

\textbf{Task-aware vs. Task-agnostic belief learning.}
Next, we compare with a design in which the belief module is trained separately from the policy, using a task-agnostic loss ($\mathcal{L}^{AR}$, Section~\ref{subsec:bm}). This echos the philosophy used in World-Models~\citep{ha2018recurrent}. As Figure~\ref{fig:perf} shows, this results in mixed success for imitation learning in POMDPs. While the agent achieves good scores in tasks such as Ant and HalfCheetah, the performance in Walker and Hopper is unsatisfactory. In contrast, BMIL, which uses a task-aware imitation-loss for the belief module, is consistently better.

\begin{figure}[h]
    \begin{center}
    \includegraphics[width=0.45\textwidth]{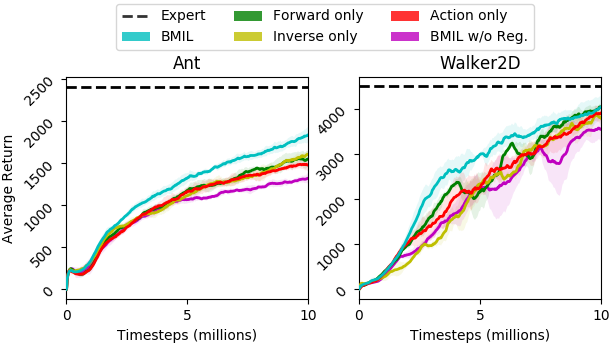}
    \caption{\small Ablation on components of belief regularization. Forward-, Inverse-, Action-only 
    correspond to using $\mathcal{L}^{f}$, $\mathcal{L}^{i}$, $\mathcal{L}^{a}$, respectively, in isolation, without the other two.}
    \label{fig:abl_reg}
    \end{center}
    \vspace{-5mm}
\end{figure}

\textbf{Are all of $\mathcal{L}^{f}, \mathcal{L}^{i}, \mathcal{L}^{a}$ useful?}
We introduced 3 different regularization terms for the belief module -- forward $(\mathcal{L}^{f})$, inverse $(\mathcal{L}^{i})$ and action $(\mathcal{L}^{a})$. To assess their benefit individually, in Figure~\ref{fig:abl_reg}, we plot learning curves for two tasks, with each of the regularizations applied in isolation. We compare them with BMIL, which includes all of them, and BMIL w/o Reg., which excludes all of them (no regularization). For the Ant task, we notice that each of $\{\mathcal{L}^{f}, \mathcal{L}^{i}, \mathcal{L}^{a}\}$ provides performance improvement over the no-regularization baseline, and combining them (BMIL) performs the best. With the Walker task, we see better mean episode-returns at the end of training with each of $\{\mathcal{L}^{f}, \mathcal{L}^{i}, \mathcal{L}^{a}\}$, compared to no-regularization; BMIL attains the best sample-complexity.

\begin{figure}[h]
    \begin{center}
    \includegraphics[width=0.45\textwidth]{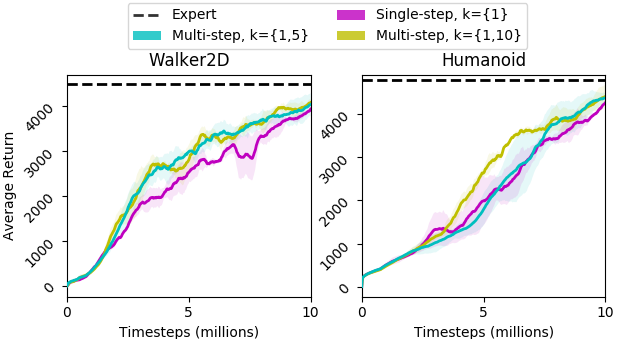}
    \caption{\small Ablation on hyperparameter $k$ in the regularization terms. Multi-step design builds over single-step by adding predictions at different temporal offsets, $k{=}5$ and $k{=}10$.}
    \label{fig:abl_k}
    \end{center}
    \vspace{-5mm}
\end{figure}

\textbf{Are multi-step predictions useful?}
As argued before, making coarse-scale, multi-step ($k{>}1$) predictions for forward, inverse observations and action-sequences could improve representations by providing temporal abstractions. In Figure~\ref{fig:abl_k}, we plot BMIL, which uses single- and multi-step losses $k{=}\{1,5\}$, and compare it with two versions: first that uses a different temporal offset $k{=}\{1,10\}$, and second that predicts only at the single-step granularity $k{=}\{1\}$. For both tasks, we get better sample-complexity and higher final mean episode-returns with multi-step, suggesting its positive contribution to representation learning.

\section{Conclusion and Future Work}
In this paper, we study imitation learning for POMDPs, which has been considerably less explored compared to imitation learning for MDPs, and learning in POMDPs with predefined reward functions. We introduce a framework comprised of a belief module, and policy and discriminator networks conditioned on the generated belief. Crucially, all networks are trained jointly with a min-max objective for adversarial imitation of expert trajectories. 

Within this flexible setup, many instantiations are possible, depending on the choice of networks. Both feed-forward and recurrent networks can be used for the policy and the discriminator, while for the belief module there is an expansive set of options based on the rich literature on representation learning, such as CPC~\citep{guo2018neural}, Z-forcing~\citep{ke2018modeling}, and using auxiliary tasks~\citep{jaderberg2016reinforcement}. Many more methods based on state-space latent-variable models are also applicable (c.f. Section~\ref{sec:rel}). In our instantiation of the belief module, we use the task-based imitation loss (Equation~\ref{eq:loss_im}), and improve robustness of representations by regularizing with multi-step prediction of past/future observations and action-sequences. One benefit of our proposed framework is that in future work, it would be straightforward to substitute other methods for learning belief representations for the one we arrived at in our paper. Similarly, recent advancements in GAN and RL literature could guide the development of better discriminator and policy networks for imitation learning in POMDPs. Exploring these ranges, as well as their interplay, are important directions.


\bibliography{uai_ref}
\bibliographystyle{uai_ref}

\clearpage
\newpage

\section{Appendix}
\begin{figure*}[t]
    \begin{center}
    \includegraphics[width=0.9\textwidth]{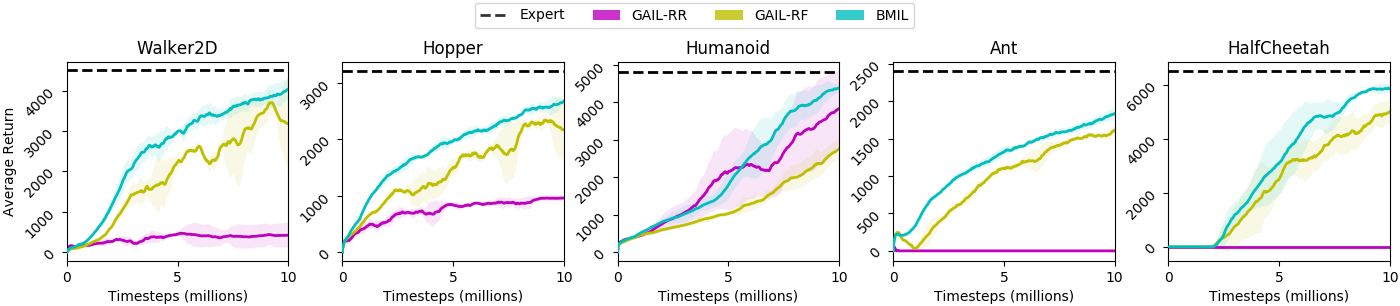}
    \caption{\small Mean episode-returns vs. timesteps of environment interaction. BMIL is our proposed architecture (Figure~\ref{fig:full_schema}); GAIL-RF uses a recurrent policy and a feed-forward discriminator, while in GAIL-RR, both the policy and the discriminator are recurrent. We plot the average and standard-deviation over 5 random seeds.}
    \label{fig:perf_gail_recurr}
    \end{center}
\end{figure*}

\subsection{Proof of inequalities in Section~\ref{sec:pol_module}}\label{appn:proofs}
We first prove the inequality connecting $D_{JS}$ between the state-visitation distribution and belief-visitation distribution of the agent and the expert:
\begin{equation*}
    D_{JS} [\rho_{\pi}(s)\ ||\ \rho_{E}(s)] \le D_{JS} [\rho_{\pi}(b)\ ||\ \rho_{E}(b)]
\end{equation*}

\begin{proof}
The proof is a simple application of the data-processing inequality for $f$-divergences~\citep{ali1966general}, of which $D_{JS}$ is a type.

We denote the filtering posterior distribution over states, given the belief, by $p(s|b)$. Note that $p(s|b)$ is characterized by the environment, and does not depend on the policy (agent or expert). The posterior over belief, given the state, however, is policy-dependent and obtained using Bayes rule as: $p_{\pi}(b|s) = \frac{p(s|b)\rho_\pi(b)}{\rho_\pi(s)}$. Also, $\rho_\pi(s,b) = \rho_\pi(s)p_\pi(b|s) = \rho_\pi(b)p(s|b)$. Analogously definitions exist for expert $E$.

We write $D_{JS} [\rho_{\pi}(b)\ ||\ \rho_{E}(b)]$ in terms of the template used for $f$-divergences. Let $f : (0, \infty) \mapsto \mathbb{R}$ be the following convex function with the property $f(1)=0$: $f(u) = -(u+1)\log \frac{1+u}{2} + u\log u$. Then,

\begin{equation*}
\begin{aligned}
D_{JS} [\rho_{\pi}(b)\ &||\ \rho_{E}(b)] \\
=& \mathbb{E}_{b \sim \rho_{E}(b)} \Big[ f(\frac{\rho_{\pi}(b)}{\rho_{E}(b)}) \Big] \\
=& \mathbb{E}_{s,b \sim \rho_{E}(s,b)} \Big[ f(\frac{\rho_{\pi}(s,b)}{\rho_{E}(s,b)}) \Big] \\ 
=& \mathbb{E}_{s \sim \rho_{E}(s)} \Big[ \mathbb{E}_{b \sim \rho_{E}(b|s)} f(\frac{\rho_{\pi}(s,b)}{\rho_{E}(s,b)}) \Big] \\ 
\text{{\em(Jensen's)}} \ge& \mathbb{E}_{s \sim \rho_{E}(s)} \Big[ f\Big(\mathbb{E}_{b \sim \rho_{E}(b|s)} \frac{\rho_{\pi}(s,b)}{\rho_{E}(s,b)}\Big) \Big] \\ 
=& \mathbb{E}_{s \sim \rho_{E}(s)} \Big[ f\Big(\mathbb{E}_{b \sim \rho_{\pi}(b|s)} \frac{\rho_{\pi}(s,b)\rho_{E}(b|s)}{\rho_{E}(s,b)\rho_{\pi}(b|s)}\Big) \Big] \\ 
=& \mathbb{E}_{s \sim \rho_{E}(s)} \Big[ f\Big(\mathbb{E}_{b \sim \rho_{\pi}(b|s)} \frac{\rho_{\pi}(s)}{\rho_{E}(s)}\Big) \Big] \\ 
=& \mathbb{E}_{s \sim \rho_{E}(s)} \Big[ f(\frac{\rho_{\pi}(s)}{\rho_{E}(s)}) \Big] \\ 
=& D_{JS} [\rho_{\pi}(s)\ ||\ \rho_{E}(s)]
\end{aligned}  
\end{equation*}
\end{proof}

Similarity, we can prove the inequality connecting $D_{JS}$ between belief-visitation distribution and belief-action-visitation distribution of the agent and the expert:
\begin{equation*}
    D_{JS} [\rho_{\pi}(b)\ ||\ \rho_{E}(b)] \le D_{JS} [\rho_{\pi}(b, a)\ ||\ \rho_{E}(b,a)]
\end{equation*}

\begin{proof}
Replace $s \mapsto b'$ and $b \mapsto (b,a)$ in the previous proof. The only {\em required} condition for that result to hold is the non-dependence of the distribution $p(s|b)$ on the policy. Therefore, if it holds that $p(b'|b,a)$ is independent of the policy, then we have, 
\begin{equation*}
    D_{JS} [\rho_{\pi}(b')\ ||\ \rho_{E}(b')] \le D_{JS} [\rho_{\pi}(b,a)\ ||\ \rho_{E}(b,a)]
\end{equation*}

The independence holds under the trivial case of a deterministic mapping $b'{=}b$. This gives us the desired inequality.
\end{proof}

\subsection{MDP and POMDP Sensors}\label{appn:sensor_info}

Description of the sensor measurements given to the agent in the MDP and POMDP environments is provided in Table~\ref{table:sensor_info}. As an example, for the Hopper task, the MDP space is 11-dimensional, which includes 6 velocity sensors and 5 position sensors; whereas the POMDP space is 5-dimensional, comprising of 5 position sensors. Amongst sensor categories, {\em velocity} includes translation and angular velocities of the torso, and also the velocities for all the joints; {\em position} includes 
torso position and orientation (quaternion), and the angle of the joints. The sensors in the MDP column marked in \textbf{bold} are removed in the POMDP setting.

\begin{table}[H]
\centering
\setlength\belowcaptionskip{-15pt}
\resizebox{\linewidth}{!}{%
\begin{tabular}{c|c|c}
\multicolumn{1}{c|}{Environment}         & \multicolumn{1}{c|}{MDP sensors $(s \in \mathcal{S})$} & \multicolumn{1}{c}{POMDP sensors $(o \in \mathcal{O})$}   \\ \hline \hline        

\multicolumn{1}{c|}{Hopper}             &
\multicolumn{1}{>{\centering\arraybackslash} p{5cm}|}{$(|\mathcal{S}|{=}11)$ \textbf{velocity(6)} + position(5)}           & \multicolumn{1}{>{\centering\arraybackslash} p{4cm}}{$(|\mathcal{O}|{=}5)$ position(5)} \\ \hline

\multicolumn{1}{c|}{Half-Cheetah}             &
\multicolumn{1}{>{\centering\arraybackslash} p{5cm}|}{$(|\mathcal{S}|{=}17)$ \textbf{velocity(9)} + position(8)}           & \multicolumn{1}{>{\centering\arraybackslash} p{4cm}}{$(|\mathcal{O}|{=}8)$ position(8)} \\ \hline

\multicolumn{1}{c|}{Walker2d}             &
\multicolumn{1}{>{\centering\arraybackslash} p{5cm}|}{$(|\mathcal{S}|{=}17)$ \textbf{velocity(9)} + position(8)}           & \multicolumn{1}{>{\centering\arraybackslash} p{4cm}}{$(|\mathcal{O}|{=}8)$ position(8)} \\ \hline

\multicolumn{1}{c|}{Inv.DoublePend.}             &
\multicolumn{1}{>{\centering\arraybackslash} p{5cm}|}{$(|\mathcal{S}|{=}11)$ \textbf{velocity(3)} + position(5) + actuator forces(3)}           & \multicolumn{1}{>{\centering\arraybackslash} p{4cm}}{$(|\mathcal{O}|{=}8)$ position(5) + actuator forces(3)} \\ \hline

\multicolumn{1}{c|}{Ant}             &
\multicolumn{1}{>{\centering\arraybackslash} p{5cm}|}{$(|\mathcal{S}|{=}111)$ \textbf{velocity(14)} + position(13) + external forces(84)}           & \multicolumn{1}{>{\centering\arraybackslash} p{4cm}}{$(|\mathcal{O}|{=}97)$ position(13) + external forces(84)} \\ \hline

\multicolumn{1}{c|}{Humanoid}             &
\multicolumn{1}{>{\centering\arraybackslash} p{5cm}|}{$(|\mathcal{S}|{=}376)$ \textbf{velocity(23) + center-of-mass based velocity(84)} + position(22) + center-of-mass based inertia(140) + actuator forces(23) + external forces(84)}           & \multicolumn{1}{>{\centering\arraybackslash} p{4cm}}{$(|\mathcal{O}|{=}269)$ position(22) + center-of-mass based inertia(140) + actuator forces(23) + external forces(84)} \\ \hline

\end{tabular}}
\caption{\small{MDP and POMDP sensors (MuJoCo). The sensors in the MDP column marked in \textbf{bold} are removed in the POMDP setting.}}
\label{table:sensor_info}
\end{table}

\subsection{Hyperparameters}\label{appn:hyperparams}

\begin{table}[H]
\centering
\setlength\belowcaptionskip{-15pt}
\resizebox{\linewidth}{!}{%
\begin{tabular}{c|c}
\multicolumn{1}{c|}{Hyper-parameter}         & \multicolumn{1}{c}{Value}    \\ \hline \hline        

\multicolumn{1}{>{\centering\arraybackslash} p{5cm}|}{Parameters for Convolution networks (encoding past \& future action-sequences)}           & \multicolumn{1}{>{\centering\arraybackslash} p{4cm}}{Layers=2, Stride=1, Padding=1, Kernel\_size=3, Num\_filters = \{5,5\}} \\ \hline

\multicolumn{1}{c|}{Belief Regularization Coefficients} & \multicolumn{1}{c}{$\lambda_1{=}\lambda_2{=}\lambda_3{=}0.2$}   \\ \hline                                                    
\multicolumn{1}{c|}{Rollout length (c) in Algorithm~\ref{algo:simple}}  & \multicolumn{1}{c}{5}    \\ \hline    

\multicolumn{1}{c|}{Size of expert demonstrations $\mathcal{M}_E$}           & \multicolumn{1}{c}{50 \text{{\small(trajectories)}}}   \\ \hline                                                     
\multicolumn{1}{c|}{Size of replay buffer $\mathcal{R}$}           & \multicolumn{1}{c}{1000  \text{{\small(trajectories)}}}     \\ \hline             
\multicolumn{1}{c|}{Optimizer, Learning Rate}       & \multicolumn{1}{c}{RMSProp, 3e-4 \text{{\small(linear-decay)}}}     \\ \hline                                                              
\multicolumn{1}{c|}{$\gamma, \lambda$ \text{{\small(GAE)}}}             & \multicolumn{1}{c}{0.99, 0.95}   \\ \hline                                              
\end{tabular}}
\end{table}

\subsection{Predictions in encoding-space}\label{appn:encoding_pred}
In our approach, we regularize with single- and multi-step predictions in the space of {\em raw observations}. For many high-dimensional, complex spaces (e.g. visual inputs), it may be more efficient to operate in a lower-dimensional, compressed encoding-space, either pre-trained, or learnt online~\citep{cuccu2019playing}. 

The encoder in our architecture (yellow box in Figure~\ref{fig:full_schema}) pre-processes the raw observations before passing them to the RNN for temporal integration. We now evaluate BMIL with single- and multi-step predictions in the space of this encoder output. For instance, the forward regularization loss function is:
\begin{equation*}
    \mathcal{L}^{f}(\phi) = \mathbb{E}_{\mathcal{R}} ||Enc(o_{t+k}) - g(b^{\phi}_{t}, a_{t:t+k-1})||^2_2 
\end{equation*}
We do not pass the gradient through the target value $Enc(o_{t+k})$. The encoder is trained online as part of the belief module. Table~\ref{table:encoding_pred} 
indicates that, for the tasks considered, BMIL performance is fairly similar when predicting in observation-space vs. encoding-space.
\begin{table}[h]
\centering
\setlength\belowcaptionskip{-15pt}
\resizebox{\linewidth}{!}{%
\begin{tabular}{c||c|c}
                         & {\vtop{\hbox{\strut BMIL: Predictions in }\hbox{\strut 
                         \:\:\:observation-space}}} &  
                         {\vtop{\hbox{\strut BMIL: Predictions in }\hbox{\strut \:\:\:encoding-space}}}
                         \\ \hline \hline
Invd.DoublePend.   & 9104 $\pm$ 134 &  8883 $\pm$ 448     \\  
Hopper      & 2665 $\pm$ 70 &   2700 $\pm$ 116    \\ 
Ant & 1832 $\pm$ 92 & 1784 $\pm$ 44     \\
Walker & 4038 $\pm$ 259 &  4043 $\pm$ 113      \\           
Humanoid   &  4382 $\pm$ 117 & 4322 $\pm$ 263        \\   
Half-cheetah  & 5860 $\pm$ 171 &  5912 $\pm$ 128       
\end{tabular}}
\caption{\small Mean and std. of episode-returns, averaged over 5 random seeds, after 10M timesteps in POMDP MuJoCo.}
\label{table:encoding_pred}
\end{table}

\subsection{Experiments in environment variants that accentuate partial observability}
\label{appn:difficult_exp}
To test robustness of BMIL, we evaluate it on two new POMDP environment variants designed to make inferring the true state from given sensors more challenging. These new environments are:

\begin{itemize}
    \item \textbf{Inv.DoublePend. from velocities only - } The partially observable {\em Inverted-Double-Pendulum} used in Section~\ref{sec:exp} removes the velocity sensors to achieve partial observability, and provides as sensors only the cart-position and sin/cos of link angles. In this new environment, we remove the previously shown sensors (cart-position and link angles), and instead provide only the velocity sensors (which were removed in our original environment). Note that the motivation is to exacerbate partial observability by restricting sensors such that inferring the true state is more challenging (i.e. it is easier to infer velocity from subsequent positions than to integrate position over time from only velocity information). Indeed, our experiments indicate this is a harder imitation learning scenario. 
    \item \textbf{Walker from velocities only - } In the same spirit as above. We remove all position sensors and instead provide only the velocity sensors to the agent.
\end{itemize}

We compare BMIL to GAIL-RF (the strongest baseline).
\begin{table}[h]
\centering
\setlength\belowcaptionskip{-15pt}
\resizebox{0.75\linewidth}{!}{%
\begin{tabular}{c||c|c}
                         & GAIL-RF &  
                         BMIL
                         \\ \hline \hline
Inv.DoublePend. \\ (velocity only)   & 4988 & 6578    \\  
Walker \\ (velocity only)   & 1539 & 4199    \\
\end{tabular}}
\caption{\small Mean episode-returns, averaged over 5 runs with random seeds, after 10M timesteps.}
\label{table:diff_envs}
\end{table}

\end{document}